\documentclass[runningheads]{llncs}
\usepackage[T1]{fontenc}
\usepackage{ulem}
\usepackage{graphicx}
\usepackage{bbding}
\usepackage{multirow}
\usepackage{graphicx}
\usepackage{booktabs}
\usepackage{amsmath}
\usepackage{amssymb}
\usepackage[table,xcdraw]{xcolor}
\usepackage{setspace}
\usepackage{hyperref}
\usepackage{orcidlink}
\hypersetup{colorlinks=true,citecolor=blue,linkcolor=blue,linktocpage=true}

\begin{document}
\title{Curriculum Prompting Foundation Models for Medical Image Segmentation}
%
\titlerunning{Curriculum Prompting Foundation Models for Medical Image Segmentation}

%
\author{Xiuqi Zheng\orcidlink{0009-0002-2814-5167} \and
Yuhang Zhang \and
Haoran Zhang\and Hongrui Liang \and Xueqi Bao \and Zhuqing Jiang\textsuperscript{\Envelope} \and Qicheng Lao\orcidlink{0000-0002-6032-8548}\textsuperscript{\Envelope}}

\authorrunning{X. Zheng et al.}
%
\institute{Beijing University of Posts and Telecommunications, Beijing, China\\
\email{qicheng.lao@bupt.edu.cn}
}

\maketitle              
\begin{abstract}
Adapting large pre-trained foundation models, e.g., SAM, for medical image segmentation remains a significant challenge. A crucial step involves the formulation of a series of specialized prompts that incorporate specific clinical instructions. Past works have been heavily reliant on a singular type of prompt for each instance, necessitating manual input of an ideally correct prompt, which is less efficient. To tackle this issue, we propose to utilize prompts of different granularity, which are sourced from original images to provide a broader scope of clinical insights. However, combining prompts of varying types can pose a challenge due to potential conflicts. In response, we have designed a coarse-to-fine mechanism, referred to as curriculum prompting, that progressively integrates prompts of different types. Through extensive experiments on three public medical datasets across various modalities, we demonstrate the effectiveness of our proposed approach, which not only automates the prompt generation process but also yields superior performance compared to other SAM-based medical image segmentation methods. Code will be available at: \href{https://github.com/AnnaZzz-zxq/Curriculum-Prompting}{https://github.com/AnnaZzz-zxq/Curriculum-Prompting}.

\end{abstract}

\section{Introduction}

Medical image segmentation is a critical area of research within medical image analysis. It plays a vital role in identifying and delineating various tissues or lesions, thereby significantly enhancing the efficiency and accuracy of medical diagnosis~\cite{cheng2022resganet}. 
Recently, with the advent of large-scale foundation models for segmentation such as SAM~\cite{kirillov2023segment}, the field of medical image segmentation has seen rapid development. SAM enables the generation of masks for regions of interest through interactive prompting, 
making it well-suited for universal medical image segmentation tasks. Several studies~\cite{he2023accuracy,putz2023segment,huang2024segment} have already explored the application of SAM in medical image segmentation. However, due to the substantial differences between natural and medical images, SAM struggles to achieve optimal segmentation performance across medical image datasets. One strategy to enhance SAM's performance in medical image segmentation involves integrating medical knowledge through specialized prompts. However, the manual generation of such prompts incurs high labor costs and yields diverse prompt quality.


To address the aforementioned challenges, this paper introduces an automated approach for identifying an optimal prompt for SAM-based medical image segmentation. Unlike conventional methods that rely on a single prompt and necessitate manual intervention, our proposed methodology leverages multiple prompt types to integrate a diverse range of image-specific details and clinical knowledge into the network. However, combining diverse knowledge domains presents a non-trivial challenge. Inspired by curriculum learning \cite{curriculumlearning}, which is motivated by the cognitive learning strategies of humans gradually acquiring knowledge from simple to complex tasks, we propose \textit{curriculum prompting}, which employs prompts that have progressively increasing granularity to systematically address segmentation challenges of varying difficulty levels, starting from coarse to fine-grained levels, to mitigate conflicts across different prompt domains. Specifically, we use mask prompts as an intermediary to gradually combine box and point prompts, refining the initial coarse mask prompt into a fine-tuned version. Unlike conventional SAM-based medical image segmentation methods that depend solely on a single prompt and necessitate the manual provision of an absolutely correct prompt, our approach significantly reduces the need for manual intervention, enabling the automatic generation of optimal prompts for SAM-based medical image segmentation based only on input medical images. In summary, our paper makes three significant contributions:

\begin{itemize}

\item[$\bullet$] Automated Prompt Generation: We propose a novel approach to automatically generate optimal prompts for SAM-based medical image segmentation, eliminating the need for manual intervention and providing more image-specific details and clinically specific knowledge to the network.

\item[$\bullet$] Curriculum Prompting Method: Our method integrates prompts of varying domains in a progressive manner, starting from coarse to fine-grained levels, which helps mitigate conflicts when simultaneously using multiple prompts from different domains.

\item[$\bullet$] Improved Segmentation Results: The combined effect of automated prompt generation and curriculum prompting leads to significantly improved segmentation results on three public medical datasets across various modalities, outperforming existing SAM-based methods qualitatively and quantitatively.

\end{itemize}


\section{Methodology}
\begin{figure}[!t]
\includegraphics[width=\textwidth]{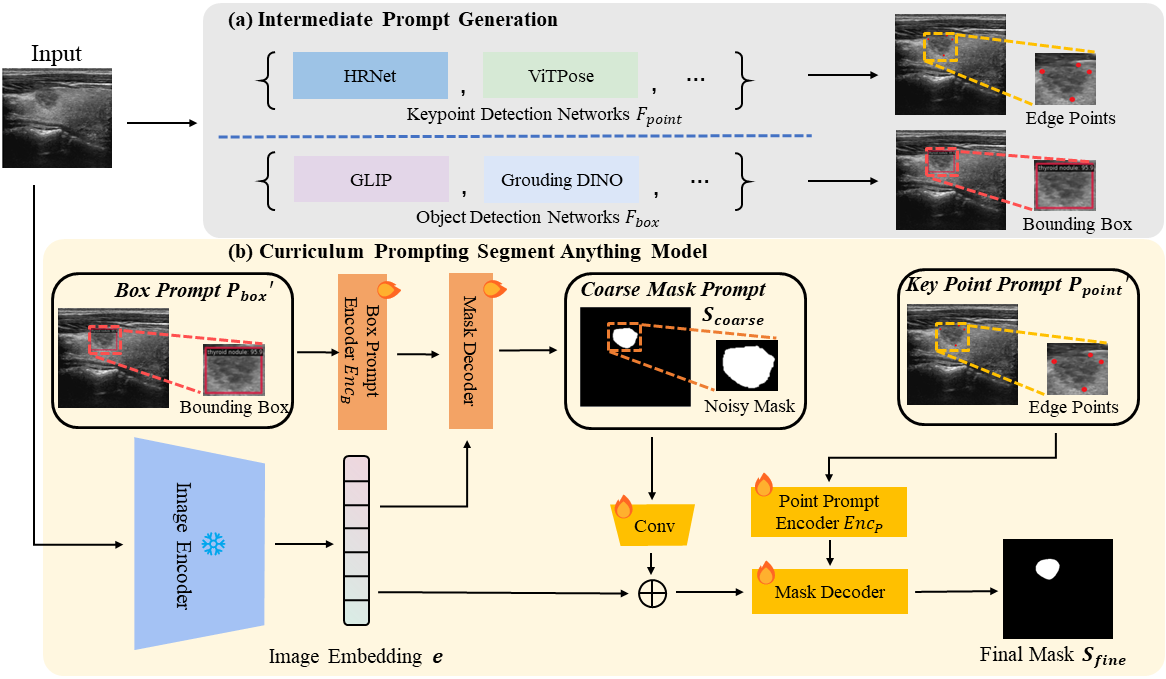}
\caption{Overview of Curriculum Prompting: (a) Intermediate Prompt Generation, which prepares prompts for SAM; (b)  Curriculum Prompting SAM,  first utilizing self-generated box prompts to obtain coarse masks, and then acquire refined masks with self-generated point prompts and coarse masks (as mask prompts).} \label{fig1}
\end{figure}

\subsection{Overview}\label{overview}

Given an image $I \in R ^{H\times W\times 3} $  with spatial resolution $H \times W$, large foundation models for segmentation, e.g., SAM, typically adopt an image encoder for extracting the image embedding $\textbf{e}$ from the image $I$, transform the prompt input $P$ through a prompt encoder $Enc$, and finally generate a segmentation mask $S$ through a mask decoder $Dec$, formulated as:
\begin{equation}
S = Dec(\textbf{e}, Enc(P)),
\end{equation}
where $P$ can be in the form of various types, such as point prompt $P_{point}=[x,y]$, where $x$, $y$ denotes the coordinates of the point, box prompt $P_{box}=[x_{1},y_{1},x_{2},y_{2}]$, composed of coordinates of the top-left and bottom-right corners of the bounding box, and mask prompt $P_{mask} \in R ^{H\times W}$. 

Prompts play a crucial role during the segmenting process, where high-quality prompts enable SAM to produce accurate segmentation masks~\cite{MIASAMformi,sam3prompts,sammed2d}. 
However, existing methods only utilize a single type of prompt, which contains limited information and often requires manual interventions.

Our proposed curriculum prompting adheres to a straightforward idea, which aims to progressively combine different types of prompts in a coarse-to-fine way. We begin with the initial prompt $P_{1}$ to assist SAM in segmentation tasks. Subsequently, the intermediate prediction generated by $P_{1}$ is fed back together with an auxiliary prompt as supplementary into SAM, initiating a recursive process. This cycle continues $n$ steps until a satisfactory segmentation result is achieved, and our empirical observations indicate that a notably improved result can be obtained when $n=2$. 
This process can be described as:
\begin{equation}
\begin{aligned}
    P_{2} &= {Dec}(\textbf{e}, {Enc}(P_{1})), \\
    P_{3} &= {Dec}(\textbf{e}, {Enc}(P_{2}, P_{2}')), \\
    &\qquad \qquad...,\\
    S &= {Dec}(\textbf{e}, {Enc}(P_{n},P_{n}')),
\end{aligned}
\end{equation}
where $P_{n}'$ denotes an auxiliary prompt apart from $P_{n}$ as a supplementary. 

In summary, we design a curriculum prompting mechanism to first address intermediate easy segmentation tasks and acquire initial coarse masks with self-generated prompts, and then add more refined prompts to tackle harder segmentation tasks and obtain the ultimate mask, to improve the overall performance.

\subsection{Coarse Prompting} \label{coarseprompting}
\noindent During the coarse prompting phase, we aim to segment most of the foreground pixels which is an easier task compared to the fine-grained segmentation with a single step. We utilize prompts that are relatively coarse but contain sufficient information to obtain an initial coarse mask. Since empirical observations suggest that two different types of prompts, e.g., box prompt and point prompt, may conflict with each other~\cite{MIASAMformi,sam3prompts}, 
in this work, we choose to employ a single type of prompt as our coarse prompt. Compared to point prompts, box prompts encompass more significant information, indicating the precise location of the object and the potential intensity features within a specified limited area. Thus, we consider self-generated box prompts as coarse prompts for initial masks.


To break through the limitation of SAM requiring manual prompts, we intend to directly and automatically derive prompts from the original image. We generate box prompts with large pre-trained object detection models, e.g. Grounding DINO~\cite{liu2023grounding} or GLIP~\cite{li2022grounded}. We fine-tune the pre-trained model with the given medical data and obtain the self-generated box prompts $P_{box}'$ as follows, 
\begin{equation}
    P_{box}' = F_{box}(I, T),
\end{equation}
where $F_{box}$ denotes the chosen object detection model, $I$ denotes the input image and $T$ denotes the text prompt if required for the model.

Following the acquisition of box prompts, a series of post-processing steps (e.g. NMS ) are undertaken. We fine-tune SAM's prompt encoder with ground-truth bounding boxes, employing a combination of Dice Loss and BCE Loss as our loss function.
Then, we acquire coarse masks $S_{coarse}$ utilizing these self-generated box prompts and the input image embedding $\textbf{e}$, 
\begin{equation}
    S_{coarse} = Dec(\textbf{e}, Enc_{B}(P_{box}')),
\end{equation}\label{equation2}
where $Enc_{B}$ denotes the prompt encoder fine-tuned with bounding boxes.

\subsection{Fine-grained Prompting}\label{Fine-grainedPrompting}

\noindent Having acquired the coarse masks, we further aim to employ more refined prompts to tackle a harder fine-grained segmentation task and guide SAM in generating the final mask. As indicated in~\cite{SAMHQ}, SAM struggles with precise edge segmentation, making the enhancement of edge delineation a more complex task compared to segmenting most of the foreground pixels. 

Thus, we adopt edge points as additional prompts to unleash SAM's full ability for segmentation. Similar to the process of box prompt generation, we employ a keypoint detection network (e.g. HRNet~\cite{sun2019deephrnet} or ViTPose~\cite{xu2022vitpose}) to generate point prompts. We obtain the self-generated point prompts $P_{point}'$ as follows:
\begin{equation} \label{eq_point}
    P_{point}' = F_{point}(I),
\end{equation}
where $F_{point}$ denotes the keypoint detection network. 


However, utilizing multiple types of prompts synergistically requires careful design. As numerous studies have indicated~\cite{MIASAMformi,0shotSAMin2Dmi,SurveyonSAMPromptEngineering}, the simultaneous use of point and box prompts can paradoxically lead to a decrease in performance. One speculation about the cause of this contradiction is due to the structure of SAM's prompt encoder. In SAM's prompt encoder $Enc$, point prompts $P_{point}$ and box prompts $P_{box}$ are processed through a series of steps and then concatenated into a sparse embedding, which is fed into the mask decoder $Dec$. During this process, different types of prompts may influence each other.

The question then arises: how can we effectively incorporate the guidance of point prompts while leveraging the information from box prompts? The answer lies in employing an additional type of prompt - the mask prompt, as a bridge to combine both box prompts and point prompts. This is where we take advantage of the coarse masks $S_{coarse}$ obtained in Section~\ref{coarseprompting}.



While point embeddings and box embeddings influence each other, the mask prompts $P_{mask}$ will only be transformed into a dense embedding through convolutions and summed with the image embedding $\textbf{e}$ without interacting with the sparse embedding. Thus, we employ self-generated point prompts $P_{point}'$ on the basis of coarse masks $S_{coarse}$ as mask prompts to achieve refined segmentation. 


Similar to the process described in Section~\ref{coarseprompting}, the SAM model we use has undergone fine-tuning with medical images, and edge points and coarse masks served as prompts. Then final masks $S_{fine}$ are acquired as follows: 

\begin{equation}
    S_{fine} = Dec(\textbf{e}, {Enc}_{P}(S_{coarse}, P_{point}')),
\end{equation}
where $Enc_{P}$ denotes the prompt encoder that is fine-tuned with edge point prompts and mask prompts, and $P_{point}'$ is obtained by Eq.(~\ref{eq_point}).

\section{Experiments and Results}
\subsection{Dataset}
\noindent We evaluate our proposed method on three public medical image datasets across various modalities, including thyroid nodule segmentation dataset TN3K~\cite{tn3k}, polyp segmentation dataset Kvasir~\cite{jha2020kvasir}, and pulmonary lesion segmentation dataset QaTa-COV19~\cite{degerli2022osegnetqata}. The TN3K dataset includes 3493 ultrasound images with pixel-wise thyroid nodule annotations; The Kvasir dataset contains 1000 endoscopic images and their corresponding polyp ground-truth masks; The QaTa-COV19 dataset consists of 9258 chest X-ray radiographs with pneumonia segmentation masks. We follow the same dataset split as ~\cite{tn3k,fan2020pranet,li2023lvit}, respectively. 

\subsection{Experiment Settings and Metrics}
\noindent Our method finetunes four distinct models, ensuring each model builds upon previous outputs. We fine-tune the object detection and keypoint detection network through the MMDetection~\cite{mmdetection} and MMPose~\cite{mmpose2020} framework. Specifically, we select Grounding DINO~\cite{liu2023grounding} and HRNet~\cite{sun2019deephrnet} for box and point prompt generation, respectively. Specifically, we use 8 edge points as point prompts. In terms of fine-tuning SAM, we initialize the model with the pre-trained weight of SAM's ViT-H version~\cite{dosovitskiy2020image}. We employ an AdamW optimizer with a learning rate of 0.0001 and a batch size of 4. Our model is implemented using PyTorch and trained and evaluated on an Nvidia RTX4090 24GB GPU. We adopt two commonly used metrics to quantitatively evaluate our proposed method, Dice (dice coefficient) and IoU (Intersection over Union). 

\subsection{Results}
\subsubsection{Our Proposed Approach Outperforms the Baselines on All Three Datasets.}
\begin{table}[!t]
\caption{Comparisons with traditional task-specific and SAM-based medical image segmentation methods. 
``*'' denotes results reported by the referenced paper. ``-'' means results are unavailable caused by dataset being used during training. }\label{tab1}
\centering
\renewcommand\arraystretch{1.2}

\resizebox{\columnwidth}{!}{

\begin{tabular}{c|cc|cc|cc}
\toprule[1pt] 
\multirow{2}{*}{Method} &
\multicolumn{2}{c|}{Kvasir (Endoscopy)} & \multicolumn{2}{c|}{TN3K (Ultrasound)} & \multicolumn{2}{c}{QaTa-COV19 (X-ray)}              \\ \cmidrule{2-7} 
                                    & mDice(\%)       & mIoU(\%)      & mDice(\%)     & mIoU(\%)     & mDice(\%)            & mIoU(\%)             \\ \midrule[1pt]
CaraNet~\cite{lou2022caranet}                   & 92.050          & 86.890        &72.647               & 62.746             &73.887 &63.517  \\
TRFE+~\cite{tn3k}                 & 42.819              & 29.517              & 83.300*       & 71.380*       & 45.719 &32.835 \\ 
LViT-T\cite{li2023lvit}                            & 77.899  & 67.519      & 76.871   & 66.573                         &77.207 &67.178 \\\midrule
fine-tuned SAM~\cite{kirillov2023segment}         & 81.848                & 74.191              &   50.791            &39.771              &48.794 &61.504 \\
nnSAM~\cite{li2023nnsam}                   &    91.176             &   85.946            & 82.797           & 74.027             & 78.943 &69.452 \\
SAM-Med2D (9 Points)~\cite{sammed2d}        &  -               & -              & 64.740              &55.760              & 76.431 & 66.083 \\
MedSAM (Box)~\cite{MedSAM}         &  86.473               &   78.046            & 81.126              &   69.464           &  - &   - \\
Grounded SAM~\cite{ren2024grounded}            & 93.340         & 89.029        & 81.600        & 73.986       &  78.625 & 68.616 \\\midrule
Ours                    & \textbf{93.670}          & \textbf{89.442}        & \textbf{84.430}        & \textbf{76.367}      &\textbf{79.826} &\textbf{70.265} \\\bottomrule[1pt] 
\end{tabular} 
}
\end{table}

We compare our method with SOTA task-specific methods and SAM-based foundation models. CaraNet~\cite{lou2022caranet}, TRFE+~\cite{tn3k} and LViT-T~\cite{li2023lvit} are three SOTA methods on the Kvasir, TN3K and QaTa-COV19 datasets, respectively. Additionally, five SOTA foundation models are chosen for comparison, including the vanilla SAM~\cite{kirillov2023segment}, nnSAM~\cite{li2023nnsam}, SAM-Med2D~\cite{sammed2d}, MedSAM~\cite{MedSAM} and Grounded SAM~\cite{ren2024grounded}. Note that we standardize the text prompt to the name or a simple description of the target lesion, such as polyp, thyroid nodule, or bilateral pulmonary infection, for fine-tuning LViT-T~\cite{li2023lvit} and Grounded SAM~\cite{ren2024grounded}. 

\begin{figure}[!t]
\includegraphics[width=\textwidth]{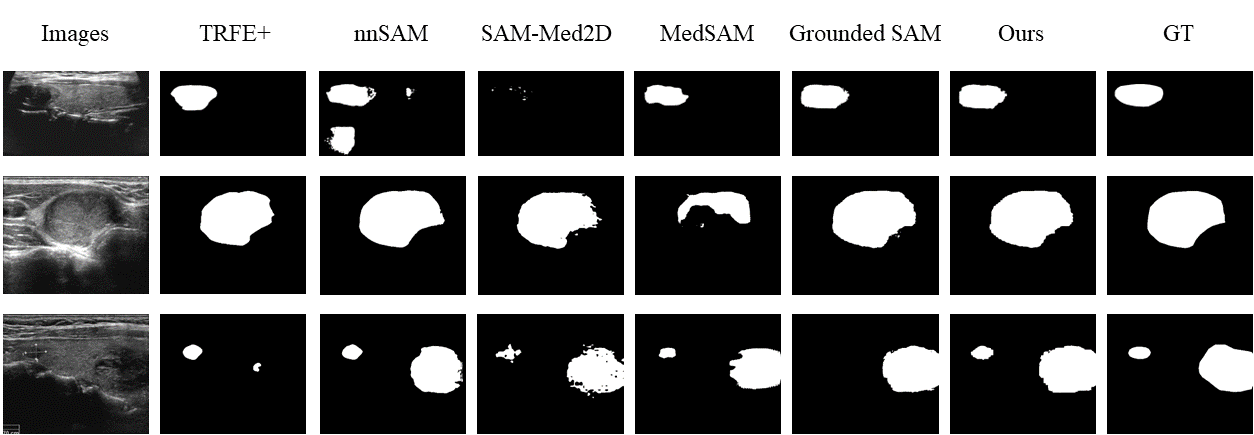}
\caption{Qualitative comparisons between our curriculum prompting SAM and other segmentation methods on the TN3K dataset, including SOTA task-specific method TRFE+, and other SAM-based segmentation models.} \label{fig2}
\end{figure}
\begin{figure}[!t]
\centering
\includegraphics[width=0.95\textwidth]{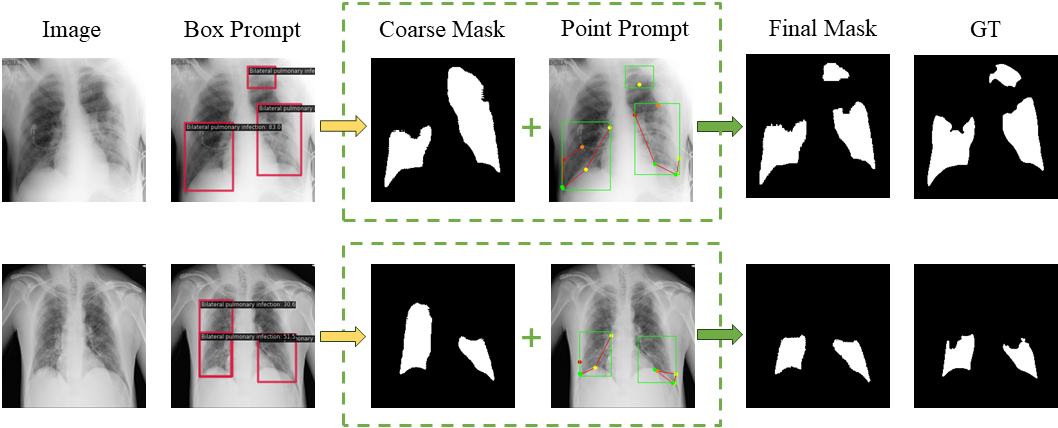}
\caption{The process of mask generation through our proposed curriculum prompting.} \label{fig3}
\end{figure}

Table.~\ref{tab1} summarizes the quantitative results. Notably, our method consistently achieves the best performance on all three tasks with average IoU scores of 89.442\%, 76.367\%, and 70.265\%. Compared to SAM-Med2D and MedSAM which require extra point prompts or box prompts derived from labels, our method outperforms them by a large margin (e.g., mean IoU > 6.9\%) without human intervention. This validates the effectiveness of our proposed method by integrating multiple prompts in a coarse-to-fine manner. 

We present qualitative results in Fig.~\ref{fig2}, where the segmentation masks of the thyroid nodules from different methods are shown. As seen in the figure, our method can precisely locate the target lesion and yields more accurate and smooth edge delineation, compared to other baselines.

\subsubsection{Visualization of Curriculum Prompting Process.}
\noindent As shown in Fig.~\ref{fig3}, during the first coarse phase when only the box prompt is used, SAM is capable of segmenting the majority of the foreground pixels. Through curriculum prompting, with the addition of edge points guidance on this basis, SAM can discern where the edges of the target are, as well as accurately distinguish between two target areas when they are nearby, instead of merging the masks into one large area. Moreover, it can be observed that the edges of the final mask have become smoother, with fewer isolated dots that are not connected to the larger area, which is very common in masks generated by SAM.

\subsubsection{Edge Points Served as Negative Prompts Can Better Improve SAM's Performance.}




\begin{table}[htbp]
  \begin{minipage}[b]{0.5\textwidth}
    \caption{Negative or positive prompts.}
\label{tab2}
    \centering
    \begin{tabular}{c|cc}
\toprule[1pt]
Label                        & Metric     & Result \\ 
\midrule[1pt]
\multirow{2}{*}{negative(0)} & mDice (\%) & \multicolumn{1}{c}{84.430} \\
                             & mIoU (\%)  & 76.367                     \\ \midrule
\multirow{2}{*}{positive(1)} & mDice (\%) & \multicolumn{1}{c}{84.259}       \\
                             &mIoU (\%)  &  76.192                          \\ \bottomrule[1pt]
\end{tabular}
  \end{minipage}
  \hfill
  \begin{minipage}[b]{0.5\textwidth}
  \caption{Ablation studies on TN3K.}
    \label{tab3}
    \centering
\resizebox{.95\textwidth}{!}{
\begin{tabular}{ccc|cc}
\toprule[1pt]
Point& BBox & Mask         & mDice (\%) & mIoU (\%) \\ \midrule[1pt]
\checkmark  &    &         &  70.300    &      61.127\\
&\checkmark      &         &       81.600&      73.986\\
& \checkmark          & \checkmark   &81.660    & 74.099     \\
\checkmark & \checkmark &           &     79.466&      71.454\\ \midrule
\checkmark   &\checkmark &   \checkmark     &       84.430&      76.367\\ \bottomrule[1pt]
\end{tabular}
}
  \end{minipage}
\end{table}



As SAM struggles with precise edge segmentation, we introduce point prompts to provide extra details, especially focusing on the lesion edges. These points can act as either positive or negative prompts. Table~\ref{tab2} demonstrates that labeling edge points as negative (label = 0) can better enhance the segmentation result. We theorize that negative prompts give more detailed guidance, clearly marking non-foreground areas. In contrast, positive prompts may not add valuable information, as the model might already identify these areas as foreground, diminishing their impact on edge definition. Thus, we label the point prompts as negative in all our experiments.

\subsubsection{Ablation Study.}



\noindent There are three different types of prompts used in our study yielding seven unique combinations. We perform ablation studies on five scenarios on the TN3K dataset, detailed in Table~\ref{tab3}. Given that mask prompts result from SAM's inference using box prompts, we exclude unavailable scenarios including solely utilizing mask prompts and utilizing both point and mask prompts due to their dependency on box prompts for mask generation. When segmenting solely with 8 edge points, SAM fails to achieve a satisfactory result, whereas, when using self-generated boxes, SAM is already capable of achieving relatively good segmentation. We can observe a decline when simultaneously using point and box prompts, compared to using box prompts alone. The results show that when utilizing three prompt types in the proposed curriculum manner, SAM gives the best segmentation performance, demonstrating each prompt type is necessary and curriculum combining them is effective.

\subsubsection{Training Time.}
\noindent The time consumption primarily occurs during the finetuning process. Our model requires 9.5h, 2.7h, 21.1h training on TN3K, Kvasir, and QaTa-COV19. For comparison, the nnSAM model takes 15.2h, 12.5h, and 20.8h. In most cases, our training time is shorter than nnSAM but outperforms nnSAM on all three datasets, demonstrating that though our training process is somewhat complicated, the training time is acceptable.

\section{Conclusion}

\noindent In this paper, we present curriculum prompting for medical image segmentation using large foundation models, an efficient method to combine multiple prompts for better segmentation performance. We employ self-generated prompts that have progressively increasing granularity to systematically address segmentation challenges of varying difficulty levels. Compared to utilizing a singular type of prompt, our method introduces more prompt information while avoiding possible conflicts between different prompt types, and achieves state-of-the-art performance on three public medical datasets with different modalities and target lesions. We hope our study provides some inspiration about prompting vision foundation models for medical image segmentation.


\begin{credits}

\subsubsection{\discintname}
The authors have no competing interests to declare that are
relevant to the content of this article.
\end{credits}
%
%

\bibliographystyle{splncs04}
\bibliography{fullpaper.bib}





\end{document}